\documentclass[fleqn,10pt]{SelfArx} 

\definecolor{color1}{RGB}{0,0,90} 
\definecolor{color2}{RGB}{0,20,20} 

\newlength{\tocsep}
\usepackage{times}
\usepackage{epsfig}
\usepackage{graphicx}
\usepackage{amsmath}
\usepackage{amssymb}
\usepackage{algorithmic}
\usepackage{algorithm}
\usepackage{memhfixc}
\usepackage{amstext}
\usepackage{cite}
\usepackage{booktabs}	

\DeclareMathOperator*{\argmax}{argmax}

\usepackage[pagebackref=true,breaklinks=true,letterpaper=true,colorlinks,bookmarks=false]{hyperref}

\JournalInfo{Concept Paper, April 2014} 
\Archive{Version~1} 

\PaperTitle{Thoughts on a Recursive Classifier Graph: \\
            a Multiclass Network for Deep Object Recognition} 

\Authors{Marius Leordeanu\textsuperscript{1}*, Rahul Sukthankar\textsuperscript{2}} 
\affiliation{\textsuperscript{1}\textit{Institute of Mathematics of the Romanian Academy}} 
\affiliation{\textsuperscript{2}\textit{Google Research and Carnegie Mellon University}} 
\affiliation{*\textbf{Corresponding author}: leordeanu@gmail.com} 

\newcommand{\keywordname}{Keywords} 

\Abstract{
We propose a general multi-class visual recognition model, termed
the Classifier Graph, which aims to generalize and integrate ideas
from many of today's successful hierarchical recognition approaches.
Our graph-based model has the advantage of enabling rich
interactions between classes from different
levels of interpretation and abstraction. The proposed multi-class
system is efficiently learned using step by step updates.  The structure consists of simple
logistic linear layers with inputs from features that are automatically selected
from a large pool. Each newly learned classifier becomes a potential new feature.
Thus, our feature pool can consist both of initial manually designed
features as well as learned classifiers from previous steps (graph nodes), each copied
many times at different scales and locations.
In this manner we can learn and grow
both a deep, complex graph of classifiers and a rich
pool of features at different levels of abstraction and interpretation.
Our proposed graph of classifiers becomes a multi-class system with
a recursive structure, suitable for deep
detection and recognition of several classes simultaneously.
}

\begin{document}

\flushbottom 

\maketitle 

\tableofcontents 

\thispagestyle{empty} 


\section{Scientific Context}
\label{sec:intro}

Visual object recognition is based on relative, hierarchical and recursive cognitive processes,
from the recognition of object parts, attributes, whole objects, interactions between them
and their contextual relationship to other objects and the scene. It is no surprise that
some of the most competitive architectures in object category recognition today
have a deep hierarchical structure~\cite{hinton_deep_learning_2006,hinton_RBM_2010}.
There are many successful hierarchical approaches,
including the face detector of Viola and Jones~\cite{ViJo04} based on classifier cascades,
the object detector of Felzenszwalb at al. with a Part-Based Model and Latent SVM's~\cite{felzenszwalb_ObjDetect_pami2010},
Conditional Random Fields~\cite{quattoni_hiddenCRFs_2007}, classification trees, random forests,
probabilistic Bayesian networks and directed acyclic graphs (DAGs)~\cite{jensen_nielsen_2007},
hierarchical hidden Markov models (HHMMs)~\cite{HMMM_1998} and methods based on feature matching with second-order
or hierarchical spatial constraints~\cite{key:leordeanu_cvpr07,Conte_GraphMatching_2004,lazebnik2006beyond}.
Even the popular nearest neighbor approach to matching
SIFT features~\cite{key:lowe}, using RANSAC with geometric verification could be modeled with a
hierarchical structure: finding correspondences between individual features
would take place at a first stage of processing,
while the rigid transformation computation and verification could be implemented with linear
transformations at higher levels of processing.
One proposal for performing such hierarchical geometric reasoning in a neural architecture is
the work by Hinton et al.\cite{hinton_capsules_2011} on \emph{capsules}.

Hierarchical classifiers are currently enjoying a great practical success due to
the development of efficient methods for deep learning in neural networks.
Considered by many to be a real scientific breakthrough in artificial intelligence,
Deep Learning has already been broadly adopted by industry in a variety of applications including
object recognition in images~\cite{Google_Image_Search_Blog, Bing_Image_Search_Blog, Baidu_Image_Search_Blog} and
speech recognition~\cite{Hinton_Acoustic_Modeling}.
Systems based on Deep Learning have won major machine learning
and computer vision competitions, such as Netflix, Kaggle and ImageNet challenges
(see~\cite{bengio_deep_learning_review_2013} for a review).
The recent success of deep classification systems and the long-term scientific interest
in their research and development strongly motivate our work on formulating a general
deep detection and recognition network with the potential to overcome many of the limitations
of current hierarchical models.

\subsection{Overview of our approach}

We propose a general recognition and learning strategy based on a
graph structure of classifiers, termed the Classifier Graph,
which aims to generalize ideas from many previous models.
The nodes of our graph are individual classifiers, which could be of any type
(e.g., anonymous intermediate classes,
object parts, attributes, objects, categories, materials or scenes).
Each classifier operates over its dedicated region, at a given location and scale relative to the image or bounding box.
It functions as a detector with a certain search area
over which it computes and returns its maximum response --- this area could be relatively small and local
or large (e.g., equal to the entire image). The classifiers at nodes in the graph influence each other through a directed set of edges,
such that the output of one (the parent) could be input to any other (the child).

The node classifiers can be viewed in the role of excitatory or inhibitory
input features, providing favorable or non-favorable \emph{context} to their children. Different from most approaches,
we make no conceptual distinction between low-level features, intermediate anonymous classes,
parts, objects, properties or context --- these are all simply classifiers and can
freely influence each other through directed links, between any two levels of abstraction.
They are free to form collectively a \emph{contextual environment} for each other.
This flexible graph structure and the classifiers at nodes
are learned from scratch over several training epochs, through an efficient supervised learning scenario combined
with a natural, unsupervised clustering and organization of the training data (Section~\ref{sec:learning}).
In a manner that is loosely reminiscent of cascade correlation~\cite{Cascade_Correlation},
each node adds a single new layer to the graph, using a logistic linear classifier with inputs
that are automatically selected from the existing pool of features (Section~\ref{sec:learning}).
The pool of features is first initialized with manually designed descriptors that operate over
raw or mid-level input (pixels, gradients, edges, color, texture, soft-segmentation etc.), randomly
sampled over many scales and
locations with different instantiating parameters.
Each new classifier learned (new node in the graph) is defined for a specific location and scale,
with a certain search area (to allow a specific location flexibility w.r.t.\ the window center of reference).
Once learned it becomes a potential new feature: copies of it at many different scales, locations
and with different search areas are added to the pool of potential features.

In this manner, we simultaneously grow both an arbitrarily complex and recursive directed graph of classifiers
and a pool of features, which represent classifiers (previously learned graphs)
at different geometric scales, levels of abstraction
and localization uncertainty.

\section{Intuition}

The graph nodes in our classifier graph
are similar to the ones in a
neural net: each one represents a logistic linear unit.
One important difference from neural networks is our ability
to connect classes from any levels of abstraction, using both
top-down as well as bottom-up links;
note that the meaning of \emph{top}
and \emph{bottom} in our hierarchy is conceptual rather than physical.
We will explain this in detail.
The edges are directed,
from an input node (this is the parent node that plays the role of an input or contextual feature)
to the classifier node (or the child). Our motivation is that parts, objects and scenes
influence each others' confidence of recognition, so
one particular class \emph{detector} could take as input the outputs of other classes' detectors.
In principle, any \emph{recognizer} at any level could function as \emph{context} or \emph{input feature}
for the recognition of any other class at any other level.

Predictions at higher levels of abstraction (\emph{I am in a room so \ldots})
could function as \emph{context} for the prediction at lower levels (\emph{\ldots I expect to see a chair}).
The probabilistic influence could also go the other way around: \emph{Since I see a chair then I could expect to be in a room}.
In the latter example, the \emph{chair} becomes context for the \emph{room}. The same two-way relationships can happen
at lower levels, between a part (e.g., a wheel) and the whole (e.g., the car).
In a classifier graph, directed edges can be formed between any two abstraction levels
in either direction: \emph{Being in a car service shop means I can expect to see car wheels} or the other way around with
a weaker but still positive influence: \emph{I see a car wheel, so I might be in a car service shop}.
In our framework, parts and objects are equal citizens: what distinguishes a part from another distinct object
or the scene is only its dedicated region (the pixels in the image corresponding to the \emph{wheel} of a car are a subset of
those that correspond to the \emph{car}; in fact, they are both \emph{car} and \emph{wheel} pixels at the same time),
whereas \emph{the mechanic}, for example, is a different object only because it does not share pixels with \emph{the car}.
Also, parts play for objects a role that is similar to the one played by objects for the scene.
Nevertheless, conceptually, all four: the part (\emph{the wheel}),
the objects (\emph{the car} and \emph{the mechanic}) and the scene (\emph{car service shop}) are just
different classes, recognized by their own dedicated classifiers, each composed of a logistic linear unit
(with its different input connections), a relative scale, location with respect to the coordinate system of the
image or bounding box, and different search area (location uncertainty over which a maximum response is computed).
Thus, our proposed system treats parts, objects, anonymous intermediate classes, materials and scenes in the same universal way:
as classifier nodes in a free graph-like structure.

\subsection{One class -- Multiple Classifiers}

Our proposal is somewhat unusual in that a single concept is represented using multiple classifiers, some of which may be learned early --- typically focusing on low-level pixel/feature inputs --- while others are learned late; the latter having many
parents that are themselves concept classifiers that aim to represent the traditional notion of parts, scene or context.  We detail this point and its consequences below.

It is clear that there is one directed edge
from the part (\emph{chair}) to the whole (\emph{room}) and a different one from the whole (\emph{room})
to the part (\emph{chair}). As parts influence the existence probability of the whole, the
presence of the whole indicates the likely presence of the part.
At the same time, it seems to lead to a
\emph{chicken and egg} problem of mutual dependencies (or cycles in a graph), which, in probabilistic inference,
is typically handled using iterative procedures.
In our case we explicitly seek a feed-forward approach, which suggests an intriguing hypothesis --- the
co-existence of several classifiers for the same class,
which act at different levels of understanding, with different triggering contextual, input
features. This design choice offers an unexpected potential benefit: that of robustness to missing
features. When one classifier for the concept lacks sufficient support, another classifier for the same class employing a
different set of features could still be sufficiently sure of its response.
Consider an example where a certain object, such as a \emph{chair}, is represented using two classifiers: 1) a classifier
based primarily upon HoG features and 2) a stronger classifier that relies on a global scene recognizer (which in turn takes
input from a variety of object classifiers, including the first chair classifier in addition to a weaker version of a scene
based on classifying Gist~\cite{Gist_Descriptor_2001} features).  Clearly, the two chair classifiers will have different failure
modes and be robust to different conditions.


Here is another example: let us imagine how
a poorer low-res independent (without input context from the outside)
\emph{person} detector could be used to increase the confidence that we are \emph{at the beach}.
That reasoning, combined with similar weak classifiers for \emph{water}, \emph{sand} and \emph{boats}
will then be used to become confident that \emph{we are indeed at the beach}. Once we \emph{see} the beach,
we will use that information to trigger more powerful classifiers (that use outside context and relationships to other classes)
for better recognizing all of the above: \emph{person, water, sand, boat and \ldots beach}, again.
Thus, by allowing multiples classifiers for the same concept we could not only
simulate any iterative inference procedure, but also move up the level of understanding, recognition and confidence.
This type of structure can be expressed very naturally in the Classifier Graph.
The idea also establishes an interesting connection to recent approaches for high-level vision tasks using
hierarchical inference with auto-context~\cite{Auto_Context_Tu_PAMI2010}.

Another advantage of multiple classifiers per class is that we could better
handle the large intra-class variability present in real-world images: people in images could appear in different
sizes, shapes and resolutions, under different poses, at different locations, in various scenes, while establishing
a wide range of interactions with other people or objects. By having many classifiers for the same category, \emph{people},
we could test all these cases simultaneously. Then, the outputs from the ensemble of \emph{person} classifiers could be
aggregated for a final answer using a variety of known methods (e.g., taking the max, weighted average or using
some voting strategy).

Here we should also mention the issue of localization.
Our classifier graph predicts the existence of the category
over a certain search area, or region of presence w.r.t.\ a center reference location,
over which a max output is returned (e.g., the classifier for
``Is there a person somewhere in the right half of the scene?'' would be different from a classifier for
``Is there a person in this precise location of the scene w.r.t.\ the image center?'').
Taking the maximum output over a search window/region is directly related to max pooling
in convolutional networks~\cite{DeepNet_ImageNet_nips2012} and maxout units from~\cite{maxout_nets_2013}.

Therefore, relative to this central location (of the image, or bounding box of attention),
we could have multiple classifiers for the same class but at different locations, scales and
with different regions of presence. We believe this idea of sharing the core of a classifier makes sense ---
let us consider the following examples: in the case of a face classifier
the right eye is different from the left eye, yet they are both \emph{eye} classifiers; for a car, the front wheels are
different from the back wheels, yet they are both ``wheels''; the person right in front of me needs a different classifier, with different properties,
than a distant person that I barely see with my peripheral vision, yet they are both ``persons'' --- I could talk to the one in front of me, but not
with the distant one. Sometimes the presence of a certain object anywhere in the scene is all we need to know, its specific location
being unimportant, while in other case the specific location is crucial: for example, during a soccer game, the location of the goalkeeper
is crucial during a penalty, but not so important when the ball is in the other half of the field --- the two cases are distinct and could
be recognized by distinct classifiers with different levels of location and pose refinement, as well as technical knowledge.

Thus, the classifier graph avoids shoehorning the multiple facets for a concept into a single classifier.  And while we believe that later classifiers are likely to contain refined versions of classifiers learned in earlier layers, this condition is not imposed --- it emerges naturally from the data when merited.

\subsection{One Graph -- Multiple Classes}

Classification at lower levels of abstraction (\emph{I see a pillow}) could help with classification at the higher levels
(\emph{It is likely to be on a bed, chair or in a closet}) and form part-whole relationships.
Classes at similar levels could influence each other
and establish ``interactions'' (\emph{If I see a hand moving in a certain familiar rhythm then I am expecting to see it touch
a pen or a keyboard}). These ideas also suggest a multi-class system in which recognition of one class is achieved
through recognition of many others --- a task that superficially appears to be a binary classification problem
(e.g., ``is there a face in this image?'') can actually be a multi-class one under the hood.  We strongly believe that since
classes are so interconnected in the real world, they should also be interconnected and jointly learned within the same
recognition system.

In the real world, related classes are likely to be found together
in a given scene and interact with each other in rich ways through
both space and time. Thus, they create a story that helps both
for recognition and in the semantic understanding of the scene.
It is a fact known since the times of the ancient Greeks,
and also confirmed by recent psychological research,
that humans are much better at learning, understanding
and remembering concepts
that are related in a coherent story~\cite{schank1995knowledge,connelly1990stories,pahl2010artifactual}.
Stories that describe interactions between objects and how they relate to each other,
both spatially and temporally, at low as well as higher levels of abstraction,
are fundamental representations for intelligent understanding.
Physical objects as well as more abstract concepts could be understood
by means of actual stories communicated verbally or experienced in real life.
Thus, the activations of classifiers
within the classifier graph can be viewed as efforts towards explaining
a complex visual scene in the form of a story.


To summarize, our multi-class system is based on the following main ideas:
\begin{itemize}
\item \textbf{Co-occurrence and interactions, object to object relationships}:
certain categories usually co-exist (co-occur).
The presence of one is strong evidence for the presence of the other.
Each could be context for the other. If two classes tend to
co-occur and interact,
then the output of a classifier for one of them would be useful as an input
feature for the classifier of the other.
\item \textbf{Relativity of classes, co-existence, opposition and
part-whole relationships:} classes sometimes exist only through their
relationship to other classes. Contrast and similarity is fundamental to ``seeing'' different classes.
\emph{White} is seen in opposition
to \emph{black}, \emph{blue} (sky) to \emph{yellow} (sun) and \emph{cold} is felt in opposition to \emph{warm}.
A flower is a flower in similarity to other flowers, but in opposition to leaves, grass or branches, and as part of trees or gardens.
To see a car we need to see at least some parts of it: the car and its
parts co-exist simultaneously, sharing the exact same 3D space. While the parts
trigger the {\it subjective, perceptual existence} of the car, the car is in fact
a different, separate entity than its parts, recognized by
at least one dedicated classifier. In human vision, the separate
existence of the car from its parts, which form it,
is also suggested by some patients suffering from
visual agnosia~\cite{farah2004visual}, who, do not seem to have a problem with
seeing, but with understanding the meaning of what they see,
or with seeing objects as {\it wholes}, which actually results
in a limited capacity to see. This brings up the question: Isn't the case
that our apparently single, unified conscious visual perception is in fact
a simultaneous multi-level process with many different subjective realities
being perceived at the same time ? Objects are also learned and understood
through stories, which give them meanings, their perception being
triggered by some behavior in a specific context:
(e.g., a big metallic looking noisy thing just passed at a high speed on the street and almost killed the curious cat).
There is a story behind every perceived thing,
and the story is often triggering the ``seeing'', the very existence of that thing in our subjective reality.
\item \textbf{Re-using prior knowledge:}
once we have learned classifiers for many categories, it would be a waste not to use them to learn new classes.
Such classifiers could include both manually designed features as well as other
feature detectors and anonymous classes discovered from previous learning tasks (e.g., using auto-encoders~\cite{hinton_deep_learning_2006,rifai2011contractive}).
Learning everything from scratch every time we deal with a new classification problem is
not efficient.
\end{itemize}

\noindent
To better appreciate the interconnectedness of classification, consider the following two examples.

\noindent \textbf{Example~1:}
Imagine that we are in the countryside and we see a horse running
through the grass.  Although we may say that we see the horse
\emph{clearly}, in reality the horse is quite far and on the basis
of shape alone, we may not be confident as to the object's identity.
However, the combination of several factors, such as its color,
the way it moves (both in terms of its articulation/gait and its
motion against the background) and its size (possibly inferred from
the nearby trees), all combine to convince us that \emph{this thing}
is a horse.  In other words, our ``seeeing'' of the horse depends a
lot on ``seeing'' as well as ``knowing'' many other things --- all
of which can be considered as inputs to our subconscious recognition
of this animal.


\noindent \textbf{Example~2:}
Consider the superficially straightforward binary classification task of recognizing a white horse against a
dark background.  How is this considered a ``multi-class'' problem in the classifier graph?

First, we note that even in the absence of external context, a horse is recognized as such through the complex interplay of various properties: its overall shape, the shapes and relative configurations of its body parts (head, four legs) and the presence of distinctive features such as a mane and tail.  Some of the body parts (e.g., eyes) could be object classifiers that are shared with broadly related animal species; others may be distinct to horses (e.g., mane) or even to this particular horse (hide pattern).  Since all of these classifiers are treated identically in the classifier graph, it is the activation patterns of all the classifiers that enables us to recognize the horse --- and just as in traditional multi-class problems, the \emph{lack} of positive firing from classifiers (e.g., wings, wheels, clothes) provides crucial information.

\begin{figure}[t!]
\begin{center}
\includegraphics[scale = 0.43, angle = 0, viewport = 0 0 350 400, clip]{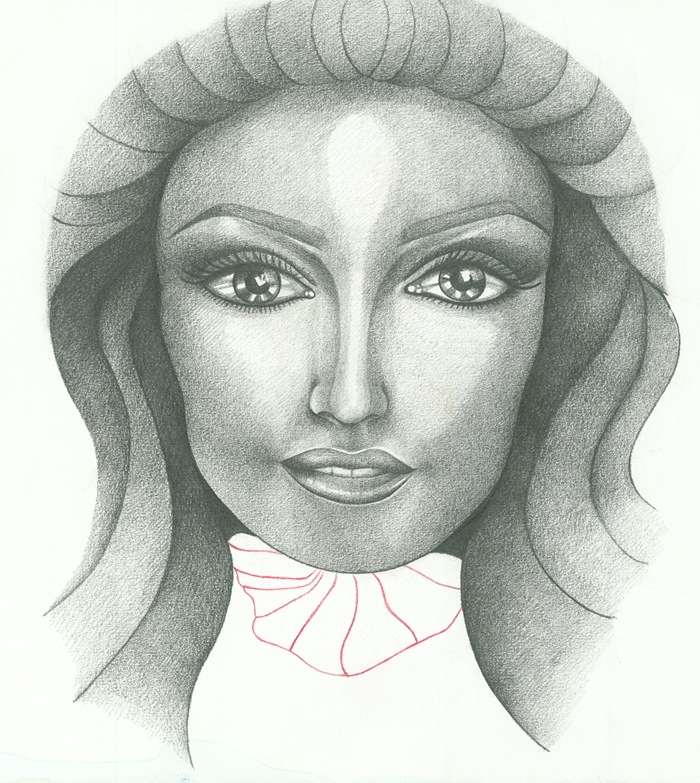}
\caption{Drawing of a woman's face. Each pixel of the drawing belongs to many
categories at the same time. For example, a pixel \emph{on the pupil}, \emph{sits} also
\emph{on an eye, a face, and a person}. Each category is \emph{seen} by different classifiers,
some specialized for the same class. Underneath the holistic visual experience of \emph{seeing a woman's face},
many classifiers at many levels of abstraction are simultaneously combined together to form
a contextual environment for each others recognition.}
\label{fig:her_eyes}
\end{center}
\end{figure}

\paragraph{Simultaneous perceptual layers of recognition:}
We have discussed the possibility of co-existence, within the structure
of the classifier graph, of several classifiers per category,
as part of a deep multi-category recognition system.
The graph is able to recognize simultaneously many
different classes and sub-classes, at different levels of abstractions,
and use them as context for each other. Classes are defined w.r.t.\ each other,
with multiple classifiers for a given class, each capturing a different individual
learning experience, from various training epochs, at different stages of
abstraction. The representation of a given class starts
with simpler, context-independent classifiers and evolves
by the addition of more complex classifiers that establish interactions to
other classifiers for objects and the scene, to eventually form
spatiotemporal stories; see Section~\ref{sec:video} for
a discussion on learning spatiotemporal categories from video.

In this section we want to pay a closer attention to the issue of simultaneous category
recognition at different interpretation levels. Let us consider the woman's face in Figure~\ref{fig:her_eyes}
and try to play a brief mind experiment.
Imagine looking at point on the pupil of the left eye of the woman's face. What type of pixel
is it? Is it a pupil pixel, an eye pixel, a face pixel or a woman's pixel? We should soon realize
that the pixel belongs to all these categories at the same time, and many more. It simultaneously sits
on a pupil, an eye, a face, and a person. Since the face image is made of pixels,
all of them have to play different roles at the same time, for a \emph{full human-like understanding} of the image.
Also note that the pixel in question could be
classified by many eye classifiers simultaneously, starting from a low-res, generic eye, to a more refined classifier
that takes in consideration more fine features of the eye. At the very top level we could have classifiers
that consider the fine geometrical alignments
to other parts of the face and be sensitive to symmetry, harmony and a general sense of beauty.

\subsection{Learning the classifier graph}
As discussed before, the output at any node in the graph could, in combination with other outputs
constitute evidence for the presence of another class. We will use
the existing classifiers at nodes
in the graph as potential features, along the initial pool of visual input features,
for learning new nodes. The idea of re-using the previously learned classifiers
as potential new input features is also related to the recent work on learning
annotations from large datasets of weakly tagged videos~\cite{aradhye2009video2text}, where Aradhye et al.\ learned annotations in stages, with each stage retaining only the most confident annotations.  Employing classifier scores from previous stages as features in the current stage enabled the system to more accurately learn annotations through this form of composition, even in the presence of label noise.

We model the classifiers at nodes with linear models, such as logistic regression,
or linear Support Vector Machines (SVM). The graph is learned node by node:
for every graph update we automatically select the relevant features (which could be previously learned nodes),
using boosting and fit a logistic linear classifier (or linear SVM).
Once the features are selected, learning is easy.
There are two difficult aspects: how to sample and organize the training set in order
to learn multi-class models and increasingly more sophisticated classifiers
for each concept, and how to choose the relevant features from the pool of
initial features and previously learned classifiers.
We employ unsupervised clustering and
boosting strategies in order to handle these issues, as discussed below.

\section{Algorithmic Formulation}

\begin{figure*}[t!]
\begin{center}
\includegraphics[scale = 0.49, angle = 0, viewport = -30 0 1000 400, clip]{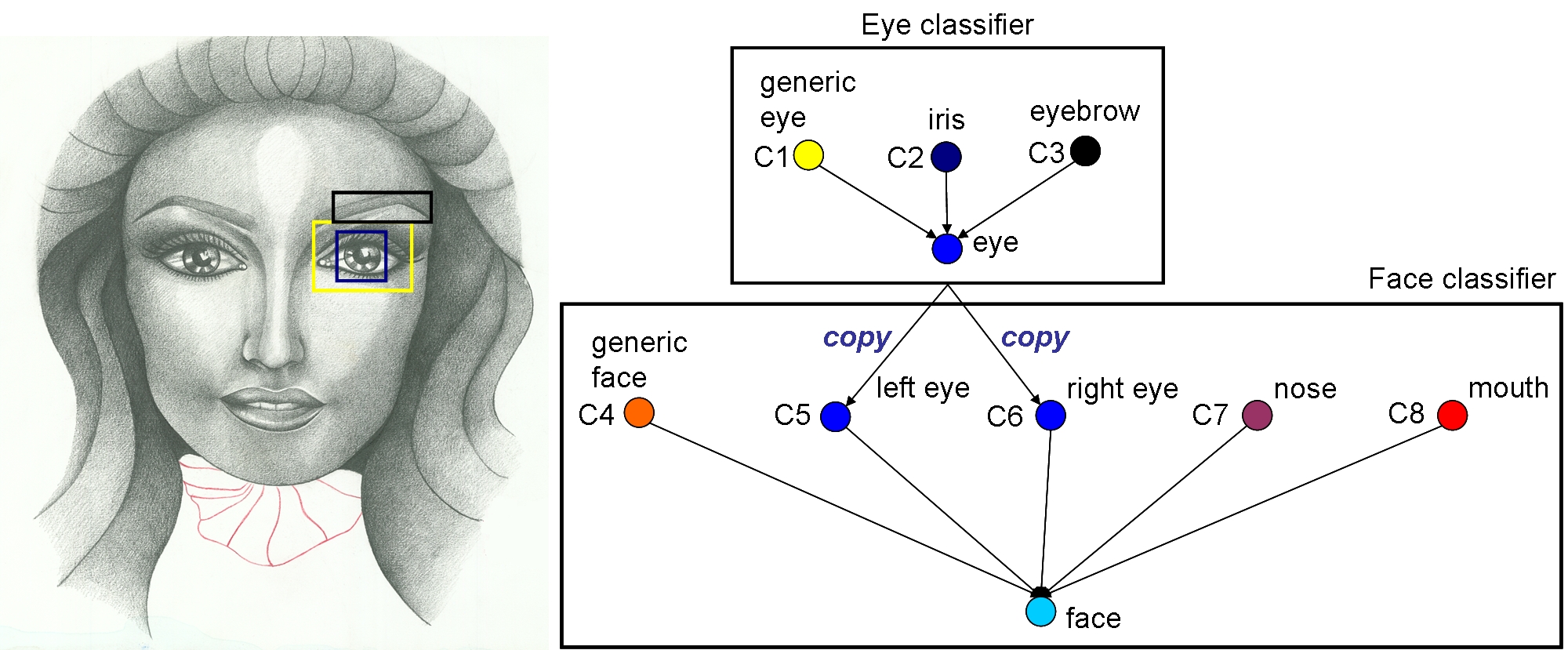}
\caption{Schematic description of a recursive, hierarchical face classifier.
The initial node classifiers use the initial features types (e.g., HoG descriptors)
and linear logistic regression. Once new classifiers (nodes) are learned, many random
copies of them at different locations, scales and with different search areas (\emph{region of presence})
are added to the pool of potential features. In this manner,
input features could achieve recursively any level of abstraction. With each learning iteration a set of
potentially more powerful and specific features are
picked automatically using boosting and a one-layer classifier is learned. The process is repeated in this
manner leading to a recursive structure, which could potentially handle any number of classes with links both
bottom-up and top-down between different levels of interpretation and abstraction. The explicit
labels used in this example (eye, nose, mouth, face) are given for clarity of presentation. In reality,
intermediate classes or sub-classes (e.g., facelets)
could be discovered and learned automatically.
In this example: an eye classifier is first learned. Copies of it are added to the pool, at different locations, scales
and with different search areas (regions of presence).
When a face classifier is learned, the eye classifier with
a given location, scale and search area, could be picked automatically and added to the graph,
together with its own parent nodes (generic eye,
iris and eyebrow) and their relative location and search area relative to their child, the {\it whole eye} classifier.
}
\label{fig:the_face_classifier}
\end{center}
\end{figure*}

The important elements in creating and growing the classifier graph
are: 1) the first features considered (the initial feature pool)
when we start learning the graph;
2) the inference method used for classification, given a certain classifier graph;
3) learning the graph: how to choose the positive and negative training examples for each epoch,
how to select the relevant features from the pool and how to learn
a new node.

\subsection{Initial Feature Types}

The initial features are the {\it atomic} classifiers that are not explained by previously trained
input nodes. They represent the first level in our hierarchical (directed graphical)
structure, most of them being rooted in the raw input from sensors.
Among the first input features, we also consider
previously trained classifiers, feature detectors and auto-encoders on various, potentially related classes.
In order to comprehensively capture different visual aspects of objects, we need to look at various
sources of information, such as shape, color, texture, occlusion regions and boundaries,
foreground/background segmentation cues. Each dimension captures a different view
of the data: the list of potential features should be comprehensive, redundant and discriminative.
Current work on deep nets spends significant computation on learning features
(sometimes referred to as basis functions or explanatory factors~\cite{bengio_deep_learning_review_2013}), directly from pixels.
The motivation behind learning directly from raw input is that hand-crafted features are not optimal.
One advantage of current neural networks systems comes from their ability to
learn specialized features at the cost of expensive computation and very large quantities of training data.
Interesting examples of learned features, besides the common-looking edges, corners or Gabor-like filters, include
neural units that encode spatial transformations~\cite{memisevic_learn_spatial_NeuralComp2010}.
Such learned features are harder to guess or design manually.

Different from the common trend of exclusively learning from data, we argue for
a more hybrid approach.
Nodes in a classifier graph may employ both engineered and learned features.
More importantly, once learned, features should be saved and re-used in future
classification tasks, instead of having to learn them from scratch for each
task. Classifier graphs can thus leverage learned
feature detectors, such as auto-encoders~\cite{hinton_deep_learning_2006,rifai2011contractive},
from previous classification tasks,
designed features, such as SIFT~\cite{key:lowe},
Shape-Context~\cite{key:belongie}
and HoG~\cite{Dalal05}, which have proved themselves in a wide range
of computer vision problems and applications, as well as learning completely new features.
Utilizing engineered features may
reduce the depth in deep, hierarchical learning, reduce both training time and
training data size, re-use prior knowledge and improve generalization.
Bootstrapping early nodes in the classifier graph with a simple HoG descriptor combined with a linear classifier may eliminate the need to learn an equivalent relatively large three-layer network that operates directly on pixels without precluding the opportunity of learning features during later stages. Thus, classifier graphs reject the false dichotomy between employing traditional engineered features or learning features from scratch in a deep architecture.

In our system, we consider, among others, some of the most
successful visual features in recognition today:
SIFT~\cite{key:lowe} (individual object matching), Haar wavelets-like
features~\cite{ViJo04} (face detection),
HoG~\cite{Dalal05} (pedestrian detection and general object category recognition)
and the similarity transformation neural units learned in~\cite{hinton_deep_learning_2006}.
We also propose local histograms over several cells of color hue values
(good for natural categories), similar local histograms of
Gabor filter responses (to encode texture and material properties), and
foreground/background segmentation cues (to capture the object's silhouette).

We extend the idea of selecting from a large pool of Haar-like features computed at many different scales and locations
(w.r.t.\ a reference box)~\cite{ViJo04} and include in our feature pool
many other classifiers and feature types, at different relative scales and locations.
We augment each classifier with a \emph{region of presence} or search area, similar to
max pooling in convolutional nets: the output returned by the feature detector is its maximum
response over the search area. We maintain a large pool of such features that is grown together with the graph.
Once learned, each new
unit node is copied many times with randomly varied scale, location and search area
and added to the feature pool for later use. This pool is effectively our overall
graph, as it contains many copies of all the subgraphs learned on the way, in a recursive
manner together with their edges. Intriguingly, the generation of copies with randomly varied
location, scale and region of presence parameters relates to the \emph{reproduction and mutation phases}
in genetic programming~\cite{koza1999genetic}, a relatively distant subfield of artificial intelligence, where the space of
computer programs is explored using Darwinian-inspired operators, such as
\emph{reproduction}, \emph{mutation} and \emph{cross-over}, that manipulate
blocks of code.
In our case, \emph{cross-over}
would correspond to combining sub-graphs of two separate
classifier nodes --- even though we did not discuss the case of creating new features by random
recombination, we do not exclude this interesting possibility for enlarging our pool of features.
Moreover, many genetic programming approaches represent programs as graph-structures, which reveals
another similarity to our approach in which classifiers are represented by graphs and sub-graphs.

The classifiers' copies are only pointers to the original classifiers,
plus the transformation parameters (location, scale and search area). We refer to the copies as
\emph{feature-nodes},
while the original learned node is termed the \emph{concept-node}, and contains the actual classifier, input links and
their weights. The classifier graph structure is deeply recursive: each concept-node is a class detector on its own,
which calls, through its input parent nodes their own concept-nodes. The recursive calls continue until
the first-stage classifiers are reached.

\subsection{Classification by Deep Detection}

Each feature-node $i$ has an associated classifier $C_i$ (a pointer to the concept-node $C_i$), a center location $p_i$ relative to its child node,
a dedicated region (scale) $S_i$ (such as a rectangle of a certain size, for a box classifier),
and a search area $A_i(p_i)$ (or region of presence) relative to its center location $p_i$. As in classical max pooling,
the classifier $C_i$ is applied at every position inside $A_i(p_i)$ and the maximum output is returned.
A concept-node is a new node learned, with no relative scale (its scale is the whole bounding box given), and no search
area (its output is considered in the middle of the bounding-box). Concept-nodes are always child nodes and all child
nodes are effectively concept-nodes (see Figure~\ref{fig:classifier_graph}).
After they are created, they are copied (as feature-nodes) and added to the feature pool for later selection
to become parent nodes.
When the node (a feature-node from the pool) has a relatively strong geometric
relationship with its child (e.g. the right eye relative to the face's center),
its area $A_i(p_i)$ of search will be small. At the other end of the spectrum,
when the node $i$ represents a completely different, location
independent object that might be anywhere in the scene,
$A_i(p_i)$ might cover the entire image (e.g., {\it Is there
a mechanic somewhere inside the car service shop?}).
The maximum classifier output,
found at each node $i$, is then passed to its children (the nodes that need it as input).
This results in a recursive algorithm for deep detection and classification (see Algorithm~\ref{alg:classification}).

\begin{algorithm}
\caption{Deep Detection: $out = DeepDetection(N_i)$}
\label{alg:classification}
\begin{algorithmic}
\STATE Goal: detect object from the level of node $N_i$.
\STATE Input: current node $N_i=\{C_i, p_i, S_i, A_i\}$.
\STATE $P \gets$ parents of node $N_i$.
\FORALL{locations $p \in A_i(p_i)$}
  \IF{$P$ is empty}
      \STATE $\mathbf{x}\gets$ feature vector at $p$ for classifier $C_i$.
  \ENDIF
  \IF{$P$ is not empty}
  \FORALL{parent nodes $N_j \in P$}
            \STATE Set location of $N_j$ relative to child: $p_j \gets p_j(p)$.
            \STATE Set response at $p_j$: $\mathbf{x}(j) = DeepDetection(N_j)$.
  \ENDFOR
  \ENDIF
  \STATE Set node response at $p$: $r(p) = C_i(\mathbf{x})$.
\ENDFOR
\STATE Max pooling over area $A_i(p_i)$:
       $out = max_{p\in A_i(p_i)}r(p)$.
\RETURN {$out$}
\end{algorithmic}
\end{algorithm}

We start from a given child node $N_{child}$ by calling the
$DeepDetection(N_{child})$ routine.
In turn, node $N_{child}$ applies the detection procedure at each of its parents nodes.
Then, they recursively call the same local detection function
at every location inside their
search area, also for
all their parents and return the maximum output. Their parents will do the same,
until the first ancestor classifier nodes are reached
(the ones rooted only in the initial features).
To avoid redundant processing on overlapping max pooling areas,
an efficient implementation of the recursion
should take advantage of dynamic
programming, caching and memoization by moving bottom up in the hierarchy and
saving the intermediate results for each search area along the way.
First, for a given child node, we can immediately find the union of
all locations, scales and search areas for all initial features and
concept-nodes that will be called during the recursive call. Starting from
the bottom-up, with the initial features first, each classifier will be called
after the concept-nodes of its parents have returned an output for
all locations and scales. We can guarantee a single call per location
and scale for a given concept-node.
The algorithm is also adaptable to parallel implementations, as independent
classifiers, for which all parents have finished, can be applied simultaneously.

\begin{figure*}[t!]
\begin{center}
\includegraphics[scale = 0.43, angle = 0, viewport = -25 0 1200 550, clip]{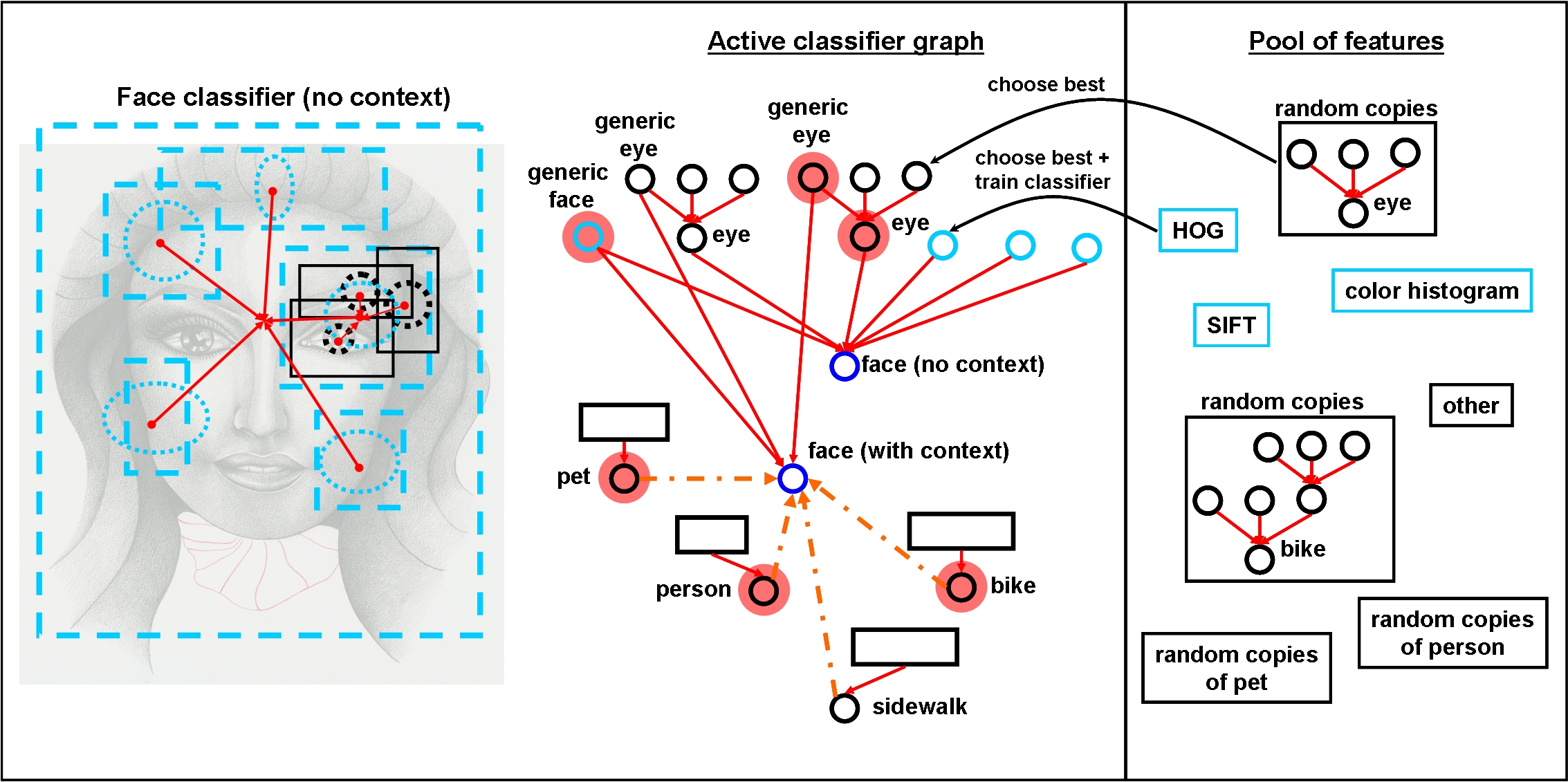}
\caption{Left: face with a superimposed context-independent part-based classifier. A separate classifier for the left
eye is chosen as an input feature for the next level face classifier. Middle: an active classifier graph with
different hierarchical levels. Red solid arrows indicate bottom-up relationships, while orange dashed-dotted arrows
show top-down or lateral relationships. Classifiers in black are subgraphs selected as input features from the pool
of features (right column). Light blue circles indicate classifiers learned on initial basic feature types
(e.g., HoG, SIFT, other existing feature detectors).
Semi-transparent red circles indicate local search areas of individual classifiers. Each area is relative to the child node.
Right column: pool of candidate input features that can contain both basic visual features as well as simple or complex learned
classifiers and their full classifier subgraphs.
}
\label{fig:the_face_classifier2}
\end{center}
\end{figure*}

Searching for the maximum output over a certain area is similar to a
scanning window strategy for detecting a single best {\it seen} object in that particular region.
As mentioned above, the search area is found relative to each of the
node's children (in the DAG): when a node has several children, its absolute
location will be different for each of them. The recursive max pooling approach
results in a hierarchical, deep detection system, also related to the deep quasi-dense
matching strategy in the recent work
on large displacement optical flow~\cite{DeepFlow_2013}.
The hierarchical classification approach follows
a part-based model, which could be arbitrarily deep and
complex when combined with the large pool of features and classifiers.
It is also suitable for
detection with contextual information, and, as mentioned before, it is meant to handle
several classifiers for the same class (see Figure~\ref{fig:the_face_classifier2}).

\section{Learning the Graph and the Pool of Features}
\label{sec:learning}

For learning the graph there are two non-trivial aspects: feature selection
and choosing the appropriate training examples for each training epoch.
For feature selection we propose a novel scheme,
weakly related to~\cite{ViJo04}, that combines the supervised Adaboost re-weighting of samples
with natural, unsupervised clustering.
During each epoch we use a certain organization of the positive training data into several, potentially overlapping,
clusters. The negative training samples, which could contain any class different from the positive label, are not clustered.
Then, at each iteration, after testing the newly added feature detector,
we apply standard Adaboost re-weighting of the training samples. The next detector selected
from the feature pool is the one with best performance on separating
the cluster with maximum sum of sample weights from the negative class.
In this manner, we take advantage of Adaboost supervised weighting to minimize the overall
ensemble exponential loss on all training samples, but we use the natural unsupervised
clustering of the positively labeled data
to select diverse feature detectors specialized for different views of the positive training set.
Interestingly enough, we have observed such a fruitful collaboration between
unsupervised clustering of the data and supervised or semi-supervised training
on two seemingly unrelated problems, that of graph matching~\cite{key:leordeanu_smoothing_optimization} and learning for graph matching~\cite{leordeanu_etal_ijcv2012}. One important aspect
here is the method(s) chosen for clustering and the distance function used
by the clustering algorithm.

\paragraph{Clustering the training data samples}
Clustering of the training data could take advantage of both {\it natural clustering} in the data
(for example, using k-means on different types of initial features and image descriptors, e.g., HoG) or/and
spatial and temporal coherence (by collecting training samples from video sequences).
Note that clusters need not be disjoint, as we choose different features and parameters for clustering,
we could end up having many clusters with overlapping elements.
Since the clusters need not be disjoint, we propose to perform several rounds of
clustering using different algorithms~\cite{kaufman2009finding} (e.g., hierarchical clustering, K-means clustering),
different descriptors of the data samples (e.g., descriptors computed at different sub-windows of
the training samples, using different types of information, such as gradients, color, word counts,
subspace and frequency analysis, just to name a few) and different distance functions,
which could range from simple Euclidean distance in feature space, to more sophisticated
distances computed from feature matching, such as the pyramid match kernel~\cite{grauman2005pyramid},
spatial pyramid kernel~\cite{lazebnik2006beyond} or matching with geometric constraints~\cite{key:lowe,key:leordeanu}.
We could keep as
final clusters the ones with high quality, estimated with both internal measures, such as the Dunn index (which
finds dense, well separated clusters) and external ones, such as cluster {\it purity}.
If negative samples are also considered
for clustering, cluster {\it purity}, in our case, would measure
the percentage of positive samples in a cluster with a positive majority.

We also expect that the spatial and temporal coherence and natural geometric and appearance
transformations, which naturally take place in video sequences, will constitute a rich source of
training data and a solid basis for clustering. Moreover, the user provided keywords on most
freely available videos on the internet, represent weak labels that could be effectively used
for grouping together videos with common labels. Such a strategy had been successfully
applied to automated video annotation on the very large video corpora from YouTube~\cite{aradhye2009video2text}.
We will come back to discuss the possibility of learning from video and representing
spatiotemporal concepts in Section~\ref{sec:video}.

Our approach has several advantages:

\begin{enumerate}
\item At each iteration, we push the next classifier to be as different as possible from the rest of the ensemble
by choosing the farthest positive cluster (in the expected error sense). This maintains classifier diversity even in our case,
when new features/classifiers are not weak. It is known that Adaboost typically does not
handle well ensembles of strong classifiers, due to lack of classifier diversity~--- strong classifiers are too good by themselves
and the different soft weighting of each sample does not help much~\cite{li_Adaboost_SVM_2008}.
\item By training over natural dense clusters we consider only representative samples and avoid over-fitting the
mislabeled or noisy data points.
\end{enumerate}

\begin{figure}[t!]
\begin{center}
\includegraphics[scale = 0.43, angle = 0, viewport = 0 0 750 450, clip]{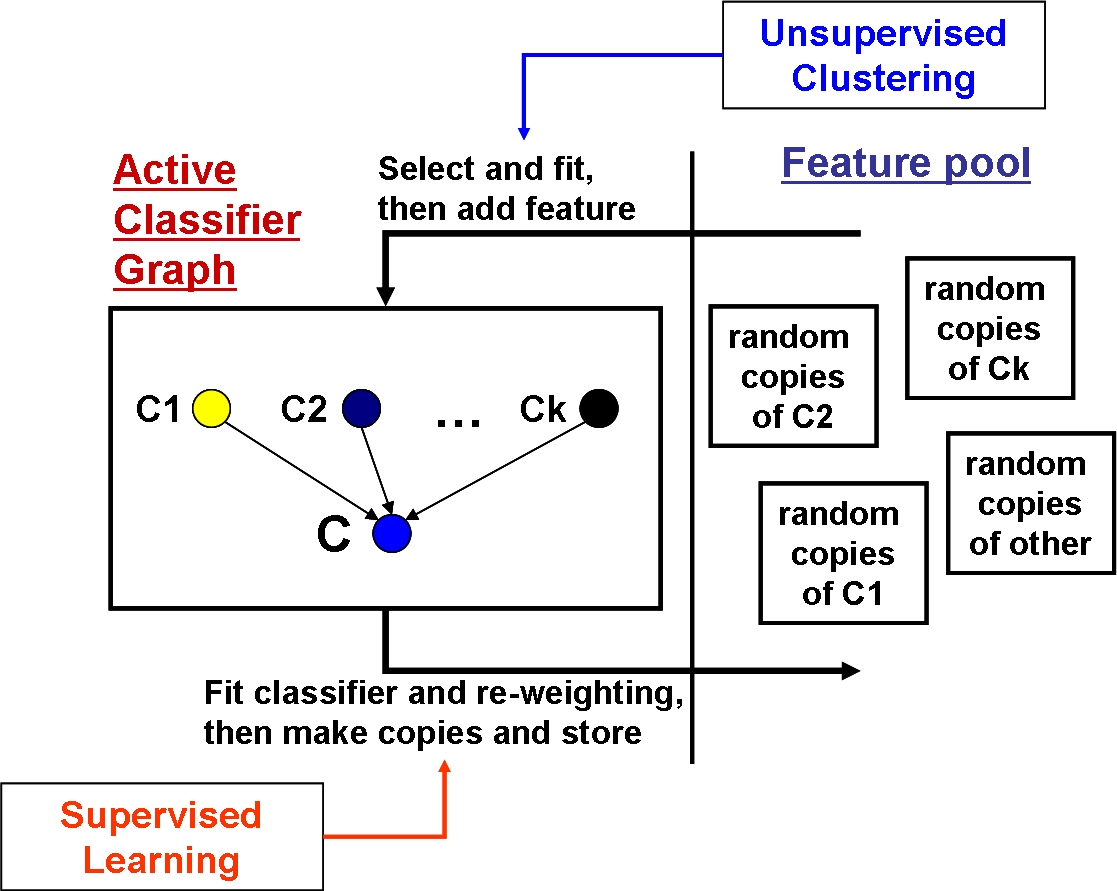}
\caption{Clusterboost: we select optimal input feature-nodes from a large pool, guided by a classification
error measure that combines unsupervised clustering with supervised weighting of the training samples.
Features are selected and added sequentially. The final classifier weights are recomputed simultaneously
with supervised learning. Many random copies of the classifier (the concept-node),
at different relative locations, scales and
with different max pooling areas are added to the feature pool, followed by improved
re-clustering using evolved distance functions that consider the new features added.
In this manner, Clusterboost can be repeated, epoch by epoch, and a large, complex classifier graph can be learned.
}
\label{fig:cluster_boost}
\end{center}
\end{figure}

\begin{figure}[t!]
\begin{center}
\includegraphics[scale = 0.4, angle = 0, viewport = 0 0 600 400, clip]{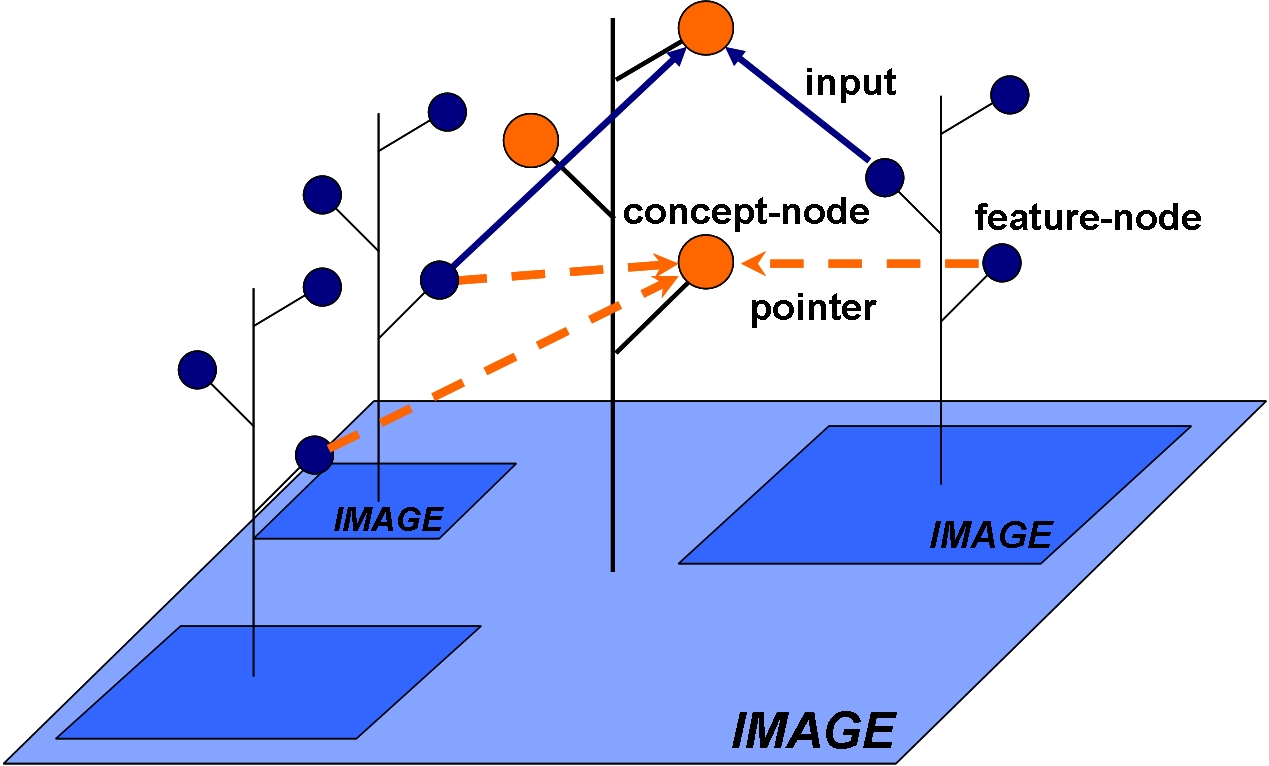}
\caption{Each new node learned is a child node, or a concept node (in orange here): it operates over the entire image (bounding box), thus it has no search area, its location is the image center and its relative scale is $1$. After it is learned it is copied many times at various locations, scales and with different search areas and added to the pool of features, as potential future parent node (here in dark blue)~--- these are the feature nodes as they link through pointers to the original classifier that they represent. A careful inspection of the figure reveals the recursive nature of classification in our graph: each parent node operates at a certain scale, or bounding box (here in dark blue), which is effectively its input image~--- it calls its classifier through a pointer and passes to it its image. From the child's point of view the image is ``the whole image'' and, in turn, the child (orange circle here) calls its parents, which again, call their orange classifier (through the pointer), and so on. The process repeats itself until orange circles with no parents are reached~--- the ones which function directly over the image input through the initial feature types. We speculate that transitions from children to parents and change in focus of attention from feature nodes to concept nodes (through pointers) may be consistent with how saccades operate in the human visual system. The feature-nodes are activated through the peripheral vision and memory system, while the orange concept nodes are activated through the attentional system.
}
\label{fig:classifier_graph}
\end{center}
\end{figure}

\noindent The overview of how we select and train new feature detectors is presented in
Algorithm~\ref{alg:cluster_boost} (Figures~\ref{fig:cluster_boost} and \ref{fig:classifier_graph}):

\begin{algorithm}
\caption{Learning with Clusterboost}
\label{alg:cluster_boost}
\begin{algorithmic}
\STATE $F$: current pool of features/classifiers
\STATE Set the observation weights $w_i = \frac{1}{N}, i=1,2,\ldots,N$.
\STATE Initialize clusters of positive samples $C_j, j = 1,2,\ldots,N_C$.
\FORALL{k = 1,2,\ldots,K}
  \STATE 1) find $C^*$ of positive samples, with maximum sum of weights $C^*=\argmax_C\sum_{i\in C}w_i$.
  \STATE 2) find best classifier $F_k(x)$ using a feature from $F$, that separates
  $C^*$ from negative data, with current weights.
  \STATE 3) let $\text{err}_k$ be weighted error of $F_k(x)$, according to $w_i's$, on all training samples.
  \STATE 4) set $\alpha_k = \log \frac{1-\text{err}_k}{\text{err}_k}$.
  \STATE 5) set $w_i \gets w_i\exp[\alpha_k I(y_i\neq F_k(x_i))]$ for all $i=1,\dots,N$.
\ENDFOR
\STATE Learn a linear logistic classifier $H$ for the current node \\ using the $K$ outputs of $F_k$ as input features.
\STATE Update feature pool $F$ with many modified copies of $H$.
\STATE End of training epoch. Return $H$ and updated $F$.
\end{algorithmic}
\end{algorithm}

First, we divide our overall training time into several epochs. During each epoch
we apply Algorithm~\ref{alg:cluster_boost}. Each epoch learns from
a different training set, based on a particular clustering of the training samples
(e.g., during each epoch we could use a specific subset of the clusters).
Regarding the training sets of each epoch, we have several possibilities:
1) for each epoch we use a different subset of the positive clusters, thus obtaining a different
{\it view} of the same class, or
2) after each epoch, we could change the label of the positive class, when in the multi-class setting.

During each epoch we limit the number of nodes to be selected with Clusterboost
at $K$; we stop early, even if classification on training set
is not perfect. That will keep the graph sparse,
avoid overfitting and add a novel node to the pool of features relatively soon.
Note that sparse networks tend to generalize better~\cite{bengio_deep_learning_review_2013},
an idea on which the recent successful Dropout training is partly based~\cite{dropout_2012}.

After selecting $K$ nodes we re-learn a logistic regression classifier, one per epoch,
to better fit the weights simultaneously (unlike Adaboost),
and obtain a probabilistic output (last part of Algorithm~\ref{alg:learning}).
After each epoch the node classifier is copied, together with a pointer to its subgraph,
at randomly varying locations $p$, scales $S$, and max pooling search areas $A$ (regions of presence),
all w.r.t.\ the reference bounding box.
These many modified copies are added to the pool of features, so that at a later epoch
we could use them as new input features.
In this manner we can have arbitrarily complex classifiers as input feature candidates, which could lead to
features of any level of abstraction and interpretation. Once a feature is selected, its subgraph (which is part of the
feature) is automatically added to the graph. The subgraph addition is not expensive, as only pointers to the subgraph
will be used in the copy of the feature.
Also, before adding a new feature from the pool we first check to see whether it is already in the graph. If a very similar
version is already there, we re-use the existing node, thus connecting nodes
across many levels of abstraction.
The overview of our overall classifier
graph learning scheme is presented in Algorithm~\ref{alg:learning}.

\begin{algorithm}
\caption{Classifier Graph Learning}
\label{alg:learning}
\begin{algorithmic}
\STATE Initialize the feature pool $F_0$.
\STATE Time starts: $t \gets 0$.
\REPEAT
  \STATE 1) A new learning epoch starts (Algorithm~\ref{alg:learning}).
  \STATE 2) Select node: train a new classifier node $H$ by Clusterboost
                  with current pool of features $F_t$.
  \STATE 3) Update graph: add $H$ and its subgraph $G_H$ to $G$.
  \STATE 4) Make random copies: generate new features $F_{new}$ of the form $\{H, p_t, S_t, A_t\}$
             by sampling from $(p,S,A)$.
  \STATE 5) Update feature pool $F_{t+1} \gets F_t \cup F_{new}$.
  \STATE 6) Select a new training set.
  \STATE 7) $t \gets t+1$. Go back to step $1$.
\UNTIL{Stopping criterion is met.}
\RETURN {Graph $G$ and feature pool $F$.}
\end{algorithmic}
\end{algorithm}

\paragraph{Evolving distance functions:}
The unsupervised clustering of the training samples strongly depends on the feature descriptors that
play an important role in defining the distance or similarity functions between training examples.
So far we have considered only the initial, manually designed feature types for building
the descriptors used for clustering. As we learn new and more powerful features we could expect to improve
our ability to perform unsupervised learning before starting a new supervised training epoch (Figure~\ref{fig:cluster_boost}).
We propose to study ways to use the classifiers learned along the way in order to better organize
the training data between training epochs. The outputs of the current classifiers could be used
to form updated descriptors of the training images. More precisely, each image $i$ could have
a descriptor vector $\mathbf{d}_i$, such that $\mathbf{d}_i(k)$ could be the output of classifier
$k$ run on the image $i$. Thus, each new classifier in the feature pool
adds a new element to the descriptor of each image. The similarity between any two images
will be a function of these descriptors and evolve from one training epoch to the next.
Consequently the unsupervised clustering will also change. This approach could
provide a natural inter-play between supervised and unsupervised learning, both
co-evolving through many stages of training. One possible similarity function could
consider the ratio of the number of co-occurring positive outputs (size of the intersection:
$\mathbf{d}_i \land \mathbf{d}_j$) to the total number of positive outputs (size of union: $\mathbf{d}_i \lor \mathbf{d}_j$).

\subsection{Human in the Loop}

The organization of the training samples could be also performed manually.
We sketch here a high-level, general strategy for possible
manual organization of the training data.
Classes and sequences initially given should be simpler:
learning basic shapes, centered, size-normalized,
with fewer colors and less cluttered backgrounds. Then, more complex scenarios
should follow: deformations, illumination changes, more difficult classes, but still relatively simple.
If possible, we should focus on categories
that are sub-parts of the final classes we want to learn, with a very consistent
and related context. Once we have the graph and the feature pool initialized
with detectors for basic categories and a relatively deep structure,
the sequences of more difficult and higher-level classes should come in.
We could first think of a specific, more limited world,
and gradually expand it. For generating different views, besides the unsupervised
clustering approach, one based on spatial and temporal coherence could also be useful (e.g., images for one view could
be from frames of the same video sequence).

While cropped images are ideal for learning bottom-up relationships, top-down relationships that go from the level of
the scene, or nearby objects, to the level of object/category of interest need information from surrounding
regions, which do not contain the object. Thus, if such contextual top-down or lateral
relationships are desired, the training images
should contain both the object of interest inside its given ground truth
bounding box/region (e.g., the eye)
as well as surrounding related areas and objects (e.g., the face, the neck, human body, other people, the whole
scene etc.). The information given for each training image
could be of similar format as in the PASCAL Challenge~\cite{everingham2010pascal},
with the ground truth bounding
boxes for all objects and categories present in the image.

\begin{figure}[t!]
\begin{center}
\includegraphics[scale = 0.52, angle = 0, viewport = -10 0 750 240, clip]{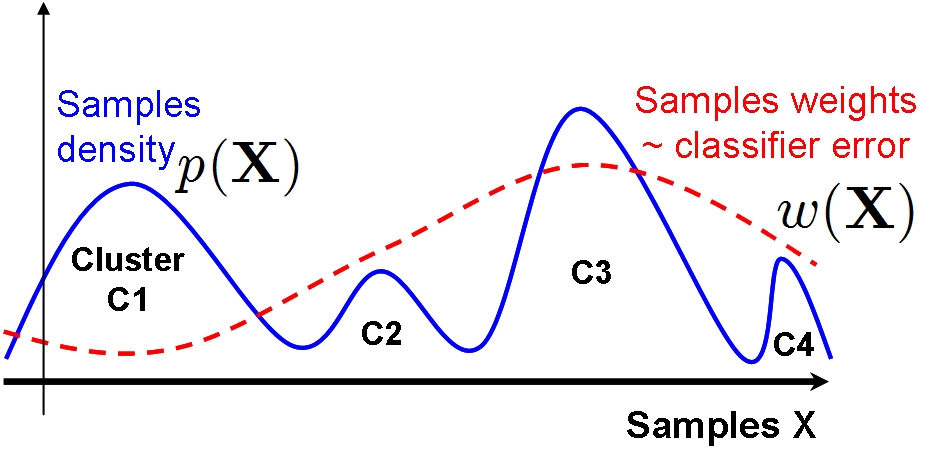}
\caption{Blue plot: natural density of the data points in some feature space. This natural
generative probability is \emph{discovered} by unsupervised clustering. The clusters are expected to be
correlated with the classification error, as points from the same cluster are similar to each other and are expected
to receive similar labels from the classification algorithm. Since supervised learning should take in consideration
the natural clustering of the data, we propose a modified, more robust version of Adaboost, which trains its weak
classifiers on individual clusters. These clusters are prioritized based on sample weights, which are correlated with
the exponential loss for Adaboost.
}
\label{fig:density_vs_error}
\end{center}
\end{figure}

\section{Implementation Details}

How large is the classifier graph, how many nodes can we expect it to have? How many edges?
What is the computational cost to perform inference once we have the graph
constructed? In order to provide approximate
answers, we first need to clarify other technical details,
such as: how many copies of a concept-node do we need to make?
Isn't this number
prohibitively large, given that we want to randomly sample locations, scales and search
areas? What does it mean to make a copy?
Given that each node is an arbitrarily complex graph, can we afford to copy it many times?

\begin{figure}[t!]
\begin{center}
\includegraphics[scale = 0.45, angle = 0, viewport = -70 0 600 400, clip]{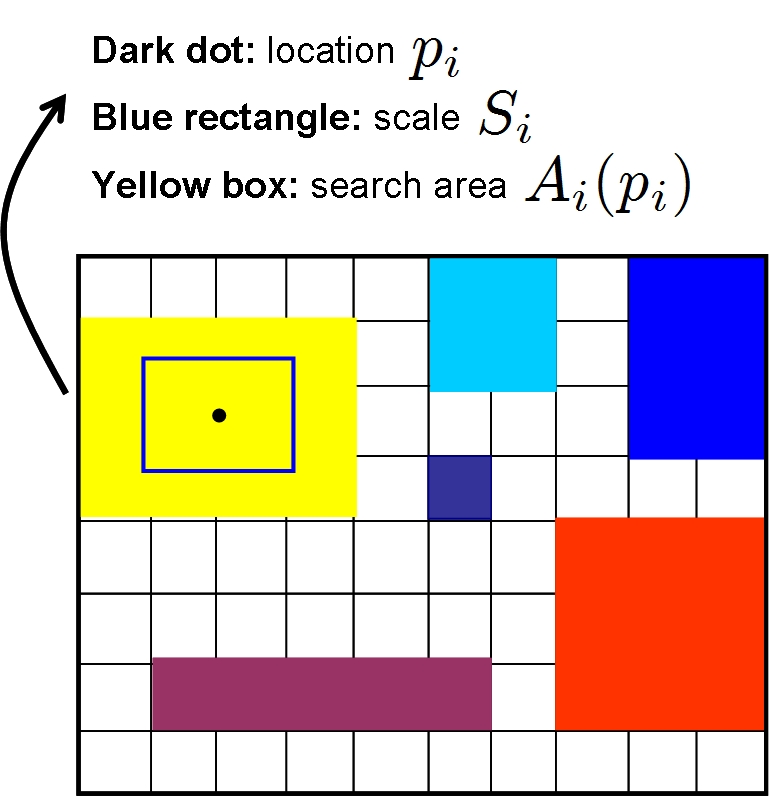}
\caption{Sampling over locations, scales and search areas cover the space uniformly.
Colored rectangles represent possible search areas for a given feature (feature-node) in the pool of features.
We argue that the granularity of sampling does not need to be very dense and that local refinements,
geometric transformations and fitting could fill the gaps in the continuum if needed.
}
\label{fig:p_S_and_A}
\end{center}
\end{figure}

Let us start with Figure~\ref{fig:p_S_and_A} and try to estimate how many copies of a node (subgraph) should
we expect to make in order to cover sufficiently dense the scale, location and search area parameters' space.
We do not expect the potential locations $p$ of objects in an image
to require very dense sampling. For example, human peripheral
vision has a very low resolution, so we do not expect us to be able to perfectly localize an object
in the image at the periphery without bringing the object into focus and performing extra
fine-tuned localization and geometric fitting, which could interpolate between discrete
locations, scales, and local changes in poses and viewpoint.
We consider that about $100$ locations would be a good rough estimate
of a sufficient number of locations.


For each location and scale we estimate, on average, about $100$ different
search areas. Of course that the actual number of locations, scales and search areas
should also depend on the actual category and could be increased or reduced after
some initial training. Buildings, for example, are large objects in most images,
they are expected to be found over a limited range of scales and locations. Birds on the
other hand are usually small and could {\it a priori} be anywhere in the image.
It would be safe to consider about $5$ different scales on average, so that
we end up with a rough estimate of
maximum $5\times10^4$ copies (feature-nodes) per classifier type.

As mentioned before, we
do not need to store an actual copy of the original child node and its subgraph (which could be as large as the whole graph),
but only a pointer to it: each of the $5\times10^4$ copies (or feature-nodes) would need only a few bytes (up to kilobytes) of information
to encode the pointer to the classifier type (the concept-node), the location, scale and search area (indexed in a finite set
of possible search areas). If we allow about $10^4$ possible
visual concepts and an average of $10$ different views (classifiers) per concept,
we end up with $10^5$ different visual classifier types (also referred to as
child nodes, or concept-nodes), for a total of about $5\times10^9$ (five billion)
feature-nodes (feature detectors) in the pool of features.
For the $10^5$ classifier types (known as child nodes or concept-nodes),
we have to store maximum $K$ input nodes (as edges/pointers)
and the weights on their incoming edges. Our initial estimate of the
storage required by the whole system would range between hundreds
of gigabytes to a few terabytes of data.

\section{Understanding video: \\ a spatiotemporal classifier graph}
\label{sec:video}

An important extension of the proposed classifier graph would be in the realm
of video sequences, for learning and recognition of spatiotemporal concepts,
such as human actions and activities, human-object interactions, and
generally, events that involve objects that act and interact over time.
Our classifier graph could be immediately extended to spatiotemporal
classes, if we see it as an {\it Always On} intelligent vision system,
which learns not only appearance-based classifiers with geometric relationships,
but also events that take place over time.
The motivation and intuition discussed in the introductory sections
apply in spatiotemporal domain as well. Scenes, objects and parts could be
statistically related not only through spatial relationships, but also through
temporal ones. A category could appear (or take place, if it is an event)
within a certain time period relative to another one, during an event of a certain type.
The time period, which relates the parent category to the child, is the time domain
equivalent of the search area, or region of presence, discussed in the spatial realm
of images. Most ideas presented in the previous sections on images, such as
multiple classes for a single category, the overall
multi-class recognition system, learning and classification,
immediately transfer in the spatiotemporal domain.
The main ideas behind the classifier graph, as also described
in the introduction, are not limited to a single moment in time.

Let us look at the following example. Imagine the category:
{\it eating a self-prepared omelette}. This is a special case
of {\it eating an omelette}, it might be one of the many classifiers
dedicated to the concept of {\it eating an omelette}, which could take
advantage of temporal context from the immediate past, when the person
who eats, had been involved in {\it preparing the omelette} - an event
that could have its own dedicated classifier. The relationships between the
two classifiers, related to eating and preparing, are temporally ordered:
preparing the food must happen before eating it. For preparing the food,
certain events should happen in a specific order: gathering the necessary
ingredients (e.g., eggs, salt, milk, oil) should be followed by
cracking the eggs into a bowl, beating them, adding a little bit of
salt and pepper, then cooking on a pan. Certain events and objects
appear in a certain temporal order and at certain relative spatial
positions, while others are less rigidly linked in space and time.
For example, while frying should definitely take place after cracking
the eggs, adding salt could, in principle, happen at any time. Exact relative
time differences and spatial dependencies might not be needed for classification,
in the same way the region of presence discussed before could vary from
very large to a single point. Some events or objects are temporally
related only by a weak co-occurrence (e.g., salt may be added at
any time during cooking a certain meal), while others are strictly ordered
in time at precise relative temporal locations (e.g., boiled eggs are ready
after three minutes of boiling).  Many people make their omelettes in different ways:
some use butter for frying, others use oil;
some add vegetables or meats, others
prefer only salt. The styles of cooking, and the manner in which the cook
performs the act of cooking will differ from person to person.
Many different classifiers might be learned for the same task of preparing
an omelette, depending on the individual experience and cultural context.

The classifier graph is a recursive network of classifiers that form
a deep graph structure. This visual system could be always on, reminding
of the classical recurrent neural networks models~\cite{williams1989learning}.
Some nodes could be triggered by previous events and maintained on for a short period of time,
while others could pay attention to present input.
Once a spatiotemporal volume is presented to a working classifier graph,
memory and attention could function together in order to perform spatiotemporal
scans and recursive recognition, by adding a third, temporal dimension
to the system presented before. A related model, the hierarchical temporal
memory system~\cite{george2005hierarchical}, also considers
temporal windows of classifier co-occurrence for establishing relationships between
discovering and recognizing spatiotemporal
patterns. It would be interested to study the connections
between the two models, in order to better understand how our
proposed classifier graph could handle the tasks and
kinds of input preferred by the hierarchical temporal
memory system.

\section{Main Contributions}

The main contributions of our classifier graph are:

\begin{enumerate}
\item We generalize the hierarchical structure of most successful methods today, by allowing
directed edges between classes at every level in the abstraction hierarchy, effectively transforming the structure
into a directed graph of classifiers. This permits a relatively simple and natural way to include
contextual information, which has proved to be a strong cue for visual recognition~\cite{Auto_Context_Tu_PAMI2010,song_context_cvpr2011,chen_context_cvpr2012,felzenszwalb_ObjDetect_pami2010}.
\item The ability to reuse resources that could be potentially useful, by maintaining a large pool
of many modified copies of old classifiers as potential input features to new classifiers.
This could lead to good generalization.
The graph, through various learning epochs, could end up learning from many different datasets, thus
collecting a varied pool of powerful features. They form
a library of classes, contexts and subparts, which are relevant to each other and to the given classification task.
We could also envision a way to {\it forget}: remove from the pool
less useful features that do not get picked.
\item Simple and effective learning at each iteration through linear logistic regression. This approach relates to recent findings in neuroscience, which show that the hippocampus (the one responsible for learning new things) learns in a similar fashion,
    one sequence of patterns  at a time~\cite{kurzweil_2012,berger2011cortical}. It also relates to efficient Deep Learning methods that either update the neural network one linear layer at a time~\cite{hinton_deep_learning_2006} or make learning
    more efficient by introducing rectified linear units --- resulting in a piecewise linear model~\cite{ReLU_icml2010}.
\item Our approach is general, uniform and recursive (the same procedure is applied for any classification task, at each iteration),
also in sync with findings that reveal the surprising uniformity of neural learning and development rules in neuroscience
~\cite{edelman1978mindful,koralek2012corticostriatal}. We offer the possibility of growing with simple and efficient
rules an arbitrarily complex, multi-classification system.
\end{enumerate}


\section*{Acknowledgments} 

\addcontentsline{toc}{section}{\hspace*{-\tocsep}Acknowledgments} 
The authors thank artist Cristina Lazar for ``Her Eyes'' --- the original
drawing of the face in Figures~\ref{fig:her_eyes}, \ref{fig:the_face_classifier} and \ref{fig:the_face_classifier2}, which has been reproduced here with the artist's permission.
The authors also thank Shumeet Baluja and Jay Yagnik for interesting discussions and valuable feedback on these ideas.
M.~Leordeanu was supported by CNCS-UEFISCDI, under PNII PCE-2012-4-0581.




\bibliographystyle{unsrt}
\bibliography{complete}


\end{document}